\documentclass{article}
\usepackage{ijcai24}

\usepackage{times}
\usepackage{soul}
\usepackage{url}
\usepackage[hidelinks]{hyperref}
\usepackage[utf8]{inputenc}
\usepackage[small]{caption}
\usepackage{graphicx}
\usepackage{amsmath}
\usepackage{amsthm}
\usepackage{booktabs}
\usepackage{algorithm}
\usepackage{algorithmic}
\usepackage[switch]{lineno}
\usepackage{amsmath}
\usepackage[inkscapelatex=false]{svg}
\usepackage{multirow}
\usepackage{amssymb}
\usepackage{url}
\usepackage{marvosym}
\usepackage{paralist}
\newcommand{\tbl}[1]{Tab. \ref{#1}}
\newcommand{\fig}[1]{Fig. \ref{#1}}
\newcommand{\eq}[1]{Eq. (\ref{#1})}
\newcommand{\ie}{\emph{i.e.}}

\title{GMC-IQA: Exploiting Global-correlation and Mean-opinion Consistency for No-reference Image Quality Assessment}
 
\author{
Zewen Chen$^{1,2}$
\and
Juan Wang$^1$\and
Bing Li$^1$ \and
Chunfeng Yuan$^{1}$ \textsuperscript{\Letter} \and
Weiming Hu $^{1,2,3}$ \and
Junxian Liu $^{4}$ \and
Peng Li $^5$ \and
Yan Wang $^5$ \and
Youqun Zhang $^5$ \and
Congxuan Zhang $^6$
\affiliations
$^1$State Key Laboratory of Multimodal Artificial Intelligence Systems, CASIA\\
$^2$School of Artificial Intelligence, University of Chinese Academy of Sciences\\
$^3$School of Information Science and Technology, ShanghaiTech University\\
$^4$Shenzhen Heytap Technology Corp., Ltd. 
$^5$Alibaba Group\\
$^6$Nanchang Hangkong University
\emails
\{chenzewen2022, jun\_wang\}@ia.ac.cn,
\{bli, cfyuan, wmhu\}@nlpr.ia.ac.cn,
James.Liu@oppo.com,
\{sanjie.lp,  wy84378, youqun.zhangyq\}@alibaba-inc.com,
zcxdsg@163.com
} 

\begin{document}

\maketitle
\begin{abstract}
Due to the subjective nature of image quality assessment (IQA), assessing which image has better quality among a sequence of images is more reliable than assigning an absolute mean opinion score for an image. Thus, IQA models are evaluated by global correlation consistency (GCC) metrics like PLCC and SROCC, rather than mean opinion consistency (MOC) metrics like MAE and MSE. However, most existing methods adopt MOC metrics to define their loss functions, due to the infeasible computation of GCC metrics during training. In this work, we construct a novel loss function and network to exploit \textbf{G}lobal-correlation and \textbf{M}ean-opinion \textbf{C}onsistency, forming a GMC-IQA framework. Specifically, we propose a novel GCC loss by defining a pairwise preference-based rank estimation to solve the non-differentiable problem of SROCC and introducing a queue mechanism to reserve previous data to approximate the global results of the whole data. Moreover,  we propose a mean-opinion network, which integrates diverse opinion features to alleviate the randomness of weight learning and enhance the model robustness. Experiments indicate that our method outperforms SOTA methods on multiple authentic datasets with higher accuracy and generalization. We also adapt the proposed loss to various networks, which brings better performance and more stable training. 
\end{abstract}

\section{Introduction}

Image quality assessment (IQA) is a long-standing research in image processing fields.
According to the availability of reference images, IQA can be categorized into three types: full-reference IQA (FR-IQA), reduced-reference IQA (RR-IQA) and no-reference IQA (NR-IQA).
Among these types, NR-IQA has gained more attention since it removes the dependence on reference images, which are unavailable in many real-world applications.

IQA is a extremely subjective task, since different people have various opinions and are hard-pressed to give the exactly same quality score for an image. Therefore, the ground truth (GT)  labels of images are defined as the average of subjective scores provided by multiple human annotators, namely mean opinion score (MOS). Compared to rating an absolute quality score for an image, it is easier and more reliable to assess the relative quality for a sequence of images. Therefore, IQA models are evaluated using Spearman's rank order correlation coefficient (SROCC) and Pearson's linear correlation coefficient (PLCC), which measure the global correlation consistency (GCC) between predicted scores and GT labels for a data set. However, most IQA methods employ distance-based loss functions, \ie MAE and MSE, for training by minimizing the discrepancy between predicted scores and GT labels to achieve the mean opinion consistency (MOC). This inconsistency between the training and evaluation objectives leads to sub-optimal performance. 

In this work, we exploit GCC and MOC for IQA from the aspect of loss function and network. To achieve the global correlation consistency, one straightforward solution is to employ the PLCC and SROCC as the training objective. However, there exist two main problems. First, the ranking function in the SROCC is non-differentiable, which makes the back-propagation of gradients infeasible. Additionally, due to the limited GPU memory, the two metric scores can only be computed within a batch during training, leading to a significant disparity compared to the results computed on the entire data set. To address these issues, we propose a pairwise preference-based rank estimation approach to build a differentiable SROCC variant. As a result, we combine the two evaluation metrics into the training objective and formulate a kind of global correlation consistency (GCC) loss. Meanwhile, to break through the limitation constrained by the GPU memory, we introduce a queue mechanism to store previous data during the training, enabling the resulting GCC loss to be approximate to that computed on the whole data set. Moreover, we theoretically prove that the proposed GCC loss can be reformulated into the MSE form by applying $\ell_2$ normalization to score values and ranking indices, which introduces strong constraints to global data correlation and global ranking consistency, respectively.

Additionally, a novel network architecture called mean-opinion network (MoNet) is proposed. Mimicking the human rating process, we develop a multi-view attention learning (MAL) module for the MoNet to implicitly learn diverse opinion features by capturing complementary contexts from various perspectives. The opinion features collected from different MALs are integrated into a comprehensive quality score, effectively relieving the impacts of hyper-parameter configurations on the performance, facilitating more robust quality score assessment. 
The MoNet is trained end to end by combining the proposed GCC loss and the MSE loss to optimize \textbf{G}lobal-correlation and \textbf{M}ean-opinion \textbf{C}onsistency, forming the GMC-IQA framework.Extensive experimental results show that the proposed GMC-IQA achieves state-of-the-art performance on multiple authentic datasets. Cross-dataset evaluation validates the better generalization of the GMC-IQA compared to many existing methods. Moreover, we experimentally prove that the proposed loss function can be adapted to various network architectures, enabling better performance and more stable training. Overall, our contributions are summarized as follows:
\begin{enumerate}
    \item We propose a novel GCC loss for IQA and develop a queue mechanism based optimization strategy, effectively addressing the unalignment between the training and evaluation objectives.
    \item We propose a mean-opinion network, which integrates diverse opinion features produced by our meticulously designed MAL modules, effectively alleviating the randomness of weight learning and enhancing the robustness of the model.
    \item Numerous experimental results validate that our GMC-IQA framework significantly outperforms many advanced methods with higher accuracy and superior generalization. We also experimentally prove the advantages of the proposed loss and network architecture in promoting the training stability and generalization. 
\end{enumerate}

\section{Related Works}

In the context of our work, we provide a brief review of related works on NR-IQA and loss functions applied in IQA.
\subsection{No-Reference Image Quality Assessment}

Due to the remarkable progress in vision applications, considerable attention has been focused on elevating the performance of IQA. As a pioneer, \cite{kang2014convolutional} design a convolutional neural network (CNN) for IQA to extract image features. Then they extend this work to a multi-task CNN \cite{kang2015simultaneous}.
However, insufficient training samples limit effective learning of CNNs-based models. 
For this reason, some methods \cite{su2020blindly,qin2023data} employ pre-trained networks, such as ResNet \cite{he2016deep} and vision transformer (ViT) \cite{dosovitskiy2020image}, as feature extractors.
However, recent research \cite{zhu2020metaiqa,chen2022teacher} point out that these popular networks pre-trained for high-level tasks are not suitable for IQA.
Therefore, some works pre-train models on related pretext tasks, \emph{e.g.}, image restoration \cite{lin2018hallucinated,ma2021blind}, quality ranking \cite{liu2017rankiqa,ma2017dipiq}, and contrastive learning \cite{zhao2023quality,madhusudana2022image}. 
Some other methods enhance the IQA performance by introducing auxiliary information. For instance, 
\cite{wang2023exploring,saha2023re} integrate textual information into the IQA.
\cite{zhang2023blind} explore the relationship among multiple tasks, namely the IQA, scene classification and distortion classification.
Additionally, many methods utilize the idea of ensemble learning to aggregate IQA-related knowledge for more effective learning. 
\cite{ma2019blind} collect a set of existing IQA models for annotation. The annotated samples are used for training their model to learn the quality score as well as the uncertainty.
Both \cite{wang2023hierarchical} and \cite{zhang2020learning} propose a novel multi-dataset training strategy.

Although existing methods have improved IQA performance by addressing various aspects of the model, they ignore to exploit the global correlation consistency in the training process, which makes a gap between training and evaluation objectives.

\subsection{Loss Functions for Image Quality Assessment}
Distance-based loss and rank-based loss are two representative losses used in IQA models.
The distance-based loss aims to minimize the difference between predicted scores and GTs. MAE and MSE are widely used for training the IQA models. When evaluating the performance of IQA models, global correlation consistency metrics, like PLCC and SROCC, are employed. This inconsistency leads to a gap between training and evaluation objectives.  
To address this problem, \cite{li2020norm} propose a Norm loss, which is formulated using a novel normalization approach. The Norm loss is closely connected to PLCC and RMSE.
It is performed within a batch during training, while PLCC and SROCC are computed on the entire test dataset during evaluation. This also contributes to inconsistent training and evaluation objectives. 

For rank-based loss, the IQA task is viewed as a quality ranking problem. 
\cite{2013Learning} adopt the cross-entropy loss to compute the discrepancy between predicted quality ranking and GT binary labels for each image pair. 
\cite{liu2017rankiqa} employ the hinge loss to formulate the optimization objective of quality ranking learning.
\cite{ma2017dipiq} use learning-to-rank algorithms, like RankNet \cite{2005Learning} and ListNet \cite{cao2007learning}, to train the IQA model on a number of image pairs. Most of these methods employ image pairs to compute the rank-based loss during training, while the evaluation is conducted on a large sequence of data to compute the ranking consistency. This discrepancy in training and evaluation can lead to sub-optimal performance.

\vspace{-0.2cm}
\section{The Proposed Method}
\subsection{Global Correlation Consistency Loss}

The performance of an IQA model is typically evaluated using two metrics: PLCC and SROCC. 
Given $n$ images, the two metrics measure the correlation consistency between the predicted scores $\mathbf{P} = [P_1, P_2, \cdots, P_n]$ and the GT labels $\mathbf{G} = [G_1, G_2, \cdots, G_n]$. They can be formulated into a unified form as follows:
\begin{equation}
\label{eq:plcc_srocc}
\rho(\mathbf{X}, \mathbf{Y})=\frac{\operatorname{cov}(\mathbf{X}, \mathbf{Y})}{\sigma_{\mathbf{X}} \sigma_{\mathbf{Y}}}.
\end{equation}
In PLCC, $\mathbf{X}=\mathbf{P}$ and $\mathbf{Y}=\mathbf{G}$; in SROCC, $\mathbf{X}=R(\mathbf{P}$) and $\mathbf{Y}=R(\mathbf{G})$, where $R(\cdot)$ is a ranking function.
In Eq. (\ref{eq:plcc_srocc}), $\sigma_{\mathbf{X}}$ and $\sigma_{\mathbf{Y}}$ denote the standard deviation of $\mathbf{X}$ and $\mathbf{Y}$, respectively, and $\operatorname{cov}(\mathbf{X}, \mathbf{Y})$ is the covariance. 

There are two main problems preventing the IQA models to be directly optimized towards the two evaluation metrics:  1) the ranking function $R(\cdot)$ in the SROCC is non-differentiable and 2) due to the limited GPU memory, the two metric scores can only be computed within a batch during training, leading to a significant disparity compared to the results computed on the entire data set.
To address these problems, we present a pairwise preference-based rank approximation approach to build a differentiable SROCC variant. Based on the PLCC and the SROCC variant, we propose a kind of global correlation consistency (GCC) loss, which enables IQA models to be optimized for the two evaluation metrics. Furthermore, we devise a queue mechanism to narrow the disparity between metric scores computed on a batch and the entire dataset.

\textbf{A) Pairwise Preference-based Rank Estimation.}
To tackle the issue of non-differentiability of the $R(\mathbf{\cdot})$, we propose a novel approach called pairwise preference-based rank estimation, which transforms the sequence ranking into pairwise comparisons. By computing the expectation of pairwise comparison results, we build a differentiable proxy function $R'(\cdot)$ to estimate the rank indices, which are statistically close to the true ranking results.

For the input data $\mathbf{P}$, we first normalize it by $S_{P_i}=(P_i - \bar{P})/\mathit{N}_{\mathbf{P}}$, where $\mathit{N}_{\mathbf{P}} = [\sum_{i=1}^n(P_i - \bar{P})^2]^{\frac{1}{2}}$ and $\bar{P}=1/n\cdot\sum_{i=1}^nP_i$. 
Then we obtain a normalized data $\mathbf{S_P} = [S_{P_1}, S_{P_2}, \cdots, S_{P_n}]$.
To get the relative comparison of any two elements in $\mathbf{S_P}$, we propose a pairwise preference label (PPL) .
The PPL $H^P_{ij}$ ($\forall i,j \in [1,n]$) is defined as the probability that $S_{p_i}$ is greater than $S_{p_j}$, which is mathematically expressed as follows:
\begin{equation}
\resizebox{.81\linewidth}{!}{$
\begin{split}
\label{eq:rank_possible}
    H^P_{ij} &= \operatorname{Pr}\left(S_{P_i}-S_{P_j}>0\right) 
    = \int_{-\infty}^{S_{P_i} - S_{P_j}} \mathbf{\omega}(s)ds \\
    &  = \int_{-\infty}^{\mathit{N}_{ij}} \frac{1}{\sqrt{2 \sigma}} e^{-\frac{s^2}{2}} ds 
     = \frac{1}{2}\left(1+\operatorname{erf}\left(\mathit{b}_{i j} / \sqrt{2} \right)\right),
\end{split}
$}
\end{equation}
where $\mathbf{\omega}(\cdot)$ denotes the probability density function of the standard normal distribution, $\mathit{b}_{ij}= S_{P_i} - S_{P_j}$ and erf$(\cdot)$ is the error function.
For any two elements $ S_{P_i}$, $S_{P_j}\in \mathbf{S_P}$, there are three possible situations: 

\noindent $\bullet$ If $S_{P_i} > S_{P_j}$, $H^P_{ij} > 0.5$
(when $S_{P_i} \gg S_{P_j}$, $H^P_{ij}\rightarrow1$); 

\noindent $\bullet$ If $S_{P_i} = S_{P_j}$, $H^P_{ij} = 0.5$;

\noindent $\bullet$ If $S_{P_i} < S_{P_j}$, $H^P_{ij} < 0.5$
(when $S_{P_i} \ll S_{P_j}$, $H^P_{ij}\rightarrow0$).

Based on Eq. (\ref{eq:rank_possible}), we obtain a probability matrix $\mathbf{H(P)}=(H^P_{ij})_{n\times n}$.
Based on the PPLs, we propose a ranking proxy function $R^\prime(\cdot)$, which estimates the rank indices for the input data $\mathbf{P}$ as follows:
\begin{equation}
\label{eq:srocc_estimate}
\resizebox{.91\linewidth}{!}{$
R^\prime(\mathbf{P})=(\sigma_{P_i})_{n\times 1}, \ \sigma_{P_i} =\mathbb{E}[\mathbf{H^P_{i\cdot}}] = \frac{1}{n} \sum_{k=1}^n H^P_{i k} ,
$}
\end{equation}
where $\sigma_{P_i}$ denotes the estimated index for the element $P_i$, and $\mathbb{E}[\cdot]$ denotes the expectation. According to Eq. (\ref{eq:srocc_estimate}), when $P_i (\forall i \in [1,n])$ is higher, the PPL $H^P_{ik} (\forall k \in [1,n])$ also tends to increase, which leads to a greater estimated index $\sigma_{P_i}$. In this manner, we achieve a statistically differentiable estimation to the non-smooth ranking function. Additionally, the proposed method preserves both monotonicity and correlation characteristics of the original ranking function.

\textbf{B) Queue Mechanism.} To address problem brought by the limited GPU memory, we propose a queue mechanism to approximate the two losses computed on a batch to those computed on the entire data set. The core idea is to maintain an on-the-fly queue for storing the latest samples. This allows us to reuse the predicted scores of the model from the immediate preceding batches. 
Specifically, the queue always maintains $K$ pairs of data, 
each of which consists of a predicted score of previous batches and the corresponding GT. 
The data in the queue are progressively replaced, where the current batch of data are enqueued and the oldest data are dequeued. 
Removing the oldest data can be beneficial, since these data are the most outdated and thus the least consistent with the newest ones, ensuring a small discrepancy between the queue data and the current batch of data. 

The queue length $K$ is an important hyper-parameter. 
The queue data always represent a sampled subset of the entire dataset. Increasing the value of $K$ can lead to a closer estimation of the PLCC and SROCC computed on the entire dataset. However, it's important to note that if $K$ is too large, it may cause the queue to retain a number of old data, potentially reducing the accuracy of ranking estimation. In the experiments, we demonstrate the impact of different values of $K$ on the performance of our model.

\textbf{C) Formulation of GCC loss.} 
Combining the pairwise preference-based rank estimation and the queue mechanism, the proposed GCC loss consists of two loss functions: a PLCC-formulated and a SGCC-formulated global correlation consistency (PGCC and SGCC) loss. They are formulated as follows:
\begin{equation}
\begin{split}
\mathcal{L}_{PGCC}(\mathbf{P}, \mathbf{G})&=1-\rho(\mathbf{P}, \mathbf{G}),\\
\mathcal{L}_{SGCC}(R'(\mathbf{P}), R'(\mathbf{G}))&=1-\rho(R'(\mathbf{P}), R'(\mathbf{G})),
\label{eq:gcc_loss}
\end{split}
\end{equation}
where $\rho$ is defined in Eq. (\ref{eq:plcc_srocc}), and $R'(\cdot)$ is  the proposed rank estimation defined in Eq. (\ref{eq:srocc_estimate}). The PGCC and the SGCC loss measure the monotonicity and the linear correlation between predicted scores and GT labels, respectively. 

\textbf{D) Theoretical Analysis.}
For the input data $\mathbf{X} = [X_1, X_2, ..., X_n]$, the normalized data $S_{X_i}$ has the property $\sum_{i=1}^nS_{X_i}^2=1$.
Thus, we can reformulate \eq{eq:plcc_srocc} to:
\begin{equation}
\label{eq:reformulate_srocc_plcc}
\centering
\resizebox{.81\linewidth}{!}{$
\begin{split}
\rho(\mathbf{X}, \mathbf{Y}) 
&= \sum_{i=1}^n\frac{X_i - \bar{X}}{\sqrt{\sum_{i=1}^n(X_i - \bar{X})^2}}\frac{Y_i - \bar{Y}}{{\sqrt{\sum_{i=1}^n(Y_i - \bar{Y})^2}}} \\
&= \sum_{i=1}^nS_{X_i}S_{Y_i} = 1 - \frac{1}{2}\sum_{i=1}^n(S_{X_i} - S_{Y_i})^2.
\end{split}
$}
\end{equation}
Then we can transform $\mathbf{\mathcal{L}_{PGCC}}$ and $\mathbf{\mathcal{L}_{SGCC}}$ to:
\begin{equation}
\label{eq:reformulate_srocc_plcc}
\resizebox{0.91\linewidth}{!}{$
\begin{split}
&\mathbf{\mathcal{L}_{PGCC}(\mathbf{P}, \mathbf{G})} = \frac{1}{2}\sum_{i=1}^n(S_{P_i} - S_{G_i})^2 = \frac{n}{2}\mathcal{L}_{MSE}(S_{P}, S_{G}), \\
&\mathbf{\mathcal{L}_{SGCC}(R^\prime(\mathbf{P}), R^\prime(\mathbf{G}))} = \frac{n}{2}\mathcal{L}_{MSE}(S_{R^\prime(\mathbf{P})}, S_{R^\prime(\mathbf{G})}).
\end{split}
$}
\end{equation}
\eq{eq:reformulate_srocc_plcc} clearly indicates that both $\mathbf{\mathcal{L}_{PGCC}}$ and $\mathbf{\mathcal{L}_{SGCC}}$ can be reformulated in the form of the MSE loss. Compared to the widely used MSE loss which computes the one-by-one distance between predicted scores and GT scores, the two losses introduce global consistency  by applying the $\ell_2$ normalization to score values and ranking indices, which impose stronger constraints on the global data distribution and ranking correlation, respectively.
\begin{figure*}[ht]
  \centering
  \includegraphics[width=\textwidth]{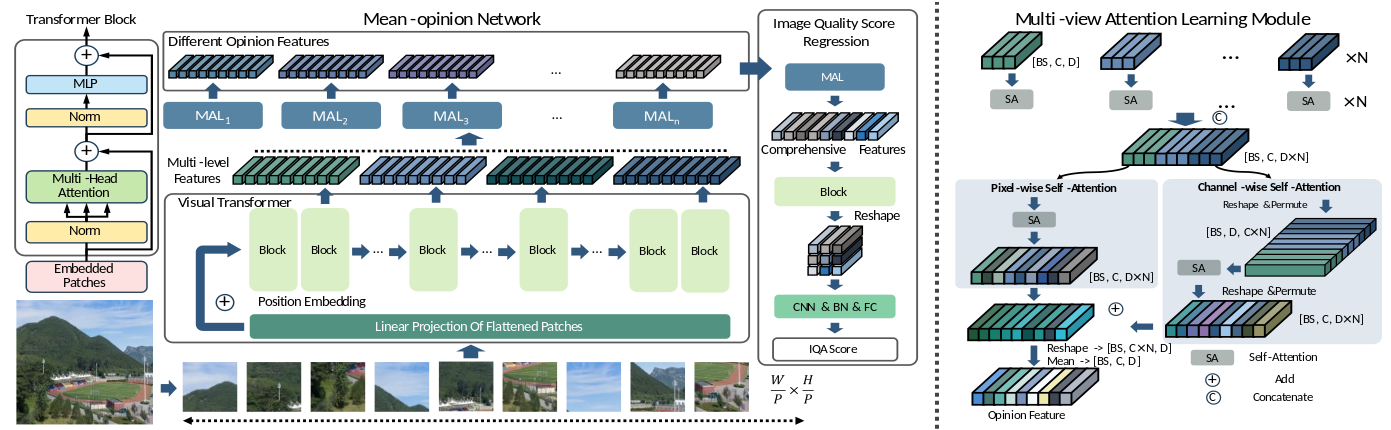}
  \caption{Network architecture of the proposed mean-opinion network (left) and multi-view attention learning module (right).}
  \label{fig:framework}
\end{figure*}

\subsection{Mean-opinion Network}
In this work, we present a novel network called mean-opinion network (MoNet), which collects various opinions by capturing diverse attention contexts to make a comprehensive decision on the image quality score. 
\fig{fig:framework} shows the network architecture of the MoNet, which mainly consists of three parts: i) a pre-trained ViT is employed for multi-level feature perception, ii) multi-view attention learning (MAL) modules are proposed for opinion collection, and iii) an image quality score regression module is designed for quality estimation. 

\textbf{A) Multi-level Semantic Perception.}
Given an image $I \in \mathbb{R}^{H \times W \times 3}$, we firstly crop it into $C$ patches with the size of $S \times S$, where $H$ and $W$ denote the height and width of the image and $C = \frac{H\times W}{S^2}$.
Then the patches are flattened and fed into a linear projection with the dimension of $D$, producing the embedding feature $\mathbf{E} \in \mathbb{R}^{C\times D}$. Subsequently, the features $\mathbf{E}$ sequentially traverses 12 transformer blocks, resulting in a set of multi-level features. Finally, the outputs of $N$ transformer blocks are selected and used as basic features, denoted as $f_j$ ($1\leq j \leq N$).

\textbf{B) Multi-view Attention Learning Module.}
The critical part of the MoNet is the multi-view attention learning (MAL) module. The motivation behind it is that individuals often have diverse subjective perceptions and regions of interest when viewing the same image. To this end, we employ multiple MALs to learn attentions from different viewpoints.
Each MAL is initialized with different weights and updated independently to encourage diversity and avoid redundant output features. The number of MALs can be flexibly set as a hyper-parameter. We show in our results its effect on the performance of our model.
As shown in \fig{fig:framework}, the MAL starts from $N$ self-attentions (SAs), each of which is responsible to process a basic feature $\mathbf{f}_j$ ($1\leq j \leq N$). 
The outputs of all the SAs are concatenated, forming a multi-level aggregated feature $\mathbf{F}\in \mathbb{R}^{C\times D \times N}$. Then $\mathbf{F}$ passes through two branches, \emph{i.e.}, a pixel-wise SA branch and a channel-wise SA branch, which apply a SA across spatial and channel dimensions, respectively, to capture complementary non-local contexts and generate multi-view attention maps.
In particular, for the channel-wise SA, the feature $\mathbf{F}$ is first reshaped and permuted to convert the size from $C\times D \times N$ to $D\times (C \times N)$. After the SA, the output feature is permuted and reshaped back to the original size $C\times D \times N$. Subsequently, the outputs of the two branches are added and average pooled, generating an opinion feature.
The design of the two branches has two key advantages. 
First, implementing the SA in different dimensions promotes diverse attention learning, yielding complementary information. Second, contextualized long-range relationships are aggregated, benefiting global quality perception.

 \textbf{C) Image Quality Score Regression.}
Assuming that $M$ opinion features are generated from $M$ MALs employed in the MoNet. To derive a global quality score from the collected opinion features, we utilize an additional MAL. The MAL integrates diverse contextual perspectives, resulting in a comprehensive opinion feature that captures essential information. This feature is then processed through a transformer block, three convolutional layers with kernel sizes of $5 \times 5$, $3 \times 3$, and $3 \times 3$ to reduce the number of channels, followed by two fully connected layers that transform the feature size from 128 to 64 and from 64 to 1. Finally, we obtain a predicted quality score from the MoNet.

\subsection{Formulation of the Training Objective}
Our total loss function combines the two GCC losses and the MSE loss. The two GCC losses are computed based on both the queue data ($\mathbf{P}^q$ and $\mathbf{G}^q$) and the current batch of data ($\mathbf{P}^b$ and $\mathbf{G}^b$), constituting a new sequence of data $\mathbf{P}^a = [\mathbf{P}^q, \mathbf{P}^b]$ and  $\mathbf{Q}^a = [\mathbf{Q}^q, \mathbf{Q}^b]$. The MSE loss is only computed on the current batch of data. 
The total loss is defined as follows:
\begin{equation}
\label{eq:final_loss}
\resizebox{.89\linewidth}{!}{$
\begin{split}
\mathcal{L}(\mathbf{P}^b, \mathbf{G}^b)=
&[\alpha\mathcal{L}_{PGCC}(\mathbf{P}^a, \mathbf{G}^a) +\beta\mathcal{L}_{SGCC}(\mathbf{P}^a, \mathbf{G}^a) \\
&+\gamma]\times \mathcal{L}_{\text{MSE}}(\mathbf{P}^b, \mathbf{G}^b),
\end{split}
$}
\end{equation}
where $\alpha$, $\beta$ and $\gamma$ are hyper-parameters. 
In \eq {eq:final_loss}, the GCC losses and the MSE loss enhance the global correlation consistency and mean opinion consistency, respectively. Therefore, the total loss is called GMC loss. When the current batch of data are easy samples, both GCC and MSE losses tend to be small, resulting in a smaller GMC loss; otherwise, the GMC loss becomes larger. Therefore, the GMC loss can pull away easy samples from hard ones, thereby strengthening the training on hard samples. We train the MoNet using the GMC loss, forming our GMC-IQA framework.

\section{Experiments}

\subsection{Datasets And Evaluation Metrics}

We train and evaluate our model on four authentic datasets, namely LIVEC \cite{ghadiyaram2015massive}, BID \cite{ciancio2010no}, Koniq \cite{hosu2020koniq}, and SPAQ \cite{fang2020perceptual}, which contain 1162, 586, 10073 and 11125 images with realistic distortions, respectively. 
PLCC and SROCC are used for evaluating the performance of IQA models. Both metrics range from -1 to 1, with higher values indicating better performance of IQA models.

\subsection{Implementation Details}
\label{subsec:Implementation}

Following the settings in  \cite{su2020blindly,yang2022maniqa,qin2023data},
we randomly divide each dataset into 80\% for training and 20\% for testing.
For the SPAQ, the shortest side of an image is resized to 512 according to \cite{fang2020perceptual}.
Let $D$ denote the number of training samples, the queue length is set to 60\%$\times D$. 
The hyper-parameters $\alpha$, $\beta$ and $\gamma$ defined in Eq. (\ref{eq:final_loss}) are set to 0.5, 0.5 and 1.
The pre-trained vit\_base\_patch8\_224 is used as the backbone of the MoNet. We use $N=4$ transformer blocks to extract basic features, namely the 3rd, 6th, 9th and 12th blocks.
If not explicitly specified, the default number of the MAL is set to $M=3$.
All experiments are conducted 10 times, and the median of the 10 scores are reported as the final score.
We use the Adam optimizer with a learning rate of $1 \times 10^{-5}$ and a weight decay of $1 \times 10^{-5}$. 
The learning rate is adjusted using the Cosine Annealing for every 50 epochs.
We train our model for 100 epochs with a batch size of 11.

\subsection{Comparisons With State-of-the-Arts}
\noindent\textbf{A) Individual Dataset Comparison.}
We compare our model with 10 NR-IQA models, namely 
CNNIQA \cite{kang2014convolutional},  
ILNIQE \cite{zhang2015feature},	 
WaDIQaM-NR \cite{bosse2017deep},	 
SFA \cite{li2018has},	 
HyperIQA \cite{su2020blindly},	 
TeacherIQA \cite{chen2022teacher}, 
MANIQA \cite{yang2022maniqa},	 
TReS \cite{golestaneh2022no},	 
DEIQT \cite{qin2023data}
 and	 
QPT-ResNet50 \cite{zhao2023quality}.

The results on the four authentic datasets are shown in \tbl{tab:individual_datasets_comparation}. The proposed GMC-IQA outperforms all the compared models on the LIVEC, BID and Koniq. 
On the SPAQ, GMC-IQA obtains a slightly inferior performance to QPT-ResNet50 \shortcite{zhao2023quality} with a margin of 0.13\% in terms of PLCC.
It is worth noting that QPT-ResNet50 synthesizes approximately $2\times10^{14}$ distorted images for pre-training. It exhibits a remarkable superiority over existing methods. Nevertheless, our GMC-IQA achieves better results to QPT-ResNet50 on most datasets without the complicated dataset synthesis for pre-training.

\begin{table*}[h]
\centering
\tiny
\resizebox{2\columnwidth}{!}{
\begin{tabular}{l|cc|cc|cc|cc}
\hline
 & \multicolumn{2}{c|}{LIVEC} & \multicolumn{2}{c|}{BID} & \multicolumn{2}{c|}{Koniq} & \multicolumn{2}{c}{SPAQ}\\
 \cline{2-9}
\multirow{-2}{*}{Model} & SROCC & PLCC & SROCC & PLCC & SROCC & PLCC & SROCC & PLCC\\ \hline
CNNIQA \cite{kang2014convolutional} & 0.6269 & 0.6008 & 0.6163 & 0.6144 & 0.6852 & 0.6837 & 0.7959 & 0.7988 \\
ILNIQE \cite{zhang2015feature} & 0.4531 & 0.5114 & 0.4946 & 0.4538 & 0.5029 & 0.4956 & 0.7194 & 0.654 \\
WaDIQaM-NR \cite{bosse2017deep} & 0.6916 & 0.7304 & 0.6526 & 0.6359 & 0.7294 & 0.7538 & 0.8397 & 0.8449\\
SFA \cite{li2018has} & 0.8037 & 0.8213 & 0.8202 & 0.8253 & 0.8882 & 0.8966 & 0.9057 & 0.9069\\
HyperIQA \cite{su2020blindly} & 0.8650 & 0.8831 & 0.8345 &	0.8757  & 0.9066 & 0.9216 & 0.9137 & 0.9170\\
TeacherIQA \cite{chen2022teacher} & 0.8411 &	0.8510  & 0.7663 &	0.7891  & 0.9100 & 0.9160 & \textbf{—} & \textbf{—} \\
MANIQA \cite{yang2022maniqa} & 0.8914&0.9089  & 0.8755 &	0.9022  & 0.9307 & 0.9452 & 0.9229 & 0.9265  \\
TReS \cite{golestaneh2022no} & 0.8427 & 0.8568 & 0.8448 & 0.8740 & 0.9263 & 0.9120  & 0.9213 & 0.9217  \\
DEIQT \cite{qin2023data} & 0.8750 & 0.8940 & \textbf{—} & \textbf{—} & 0.9210 & 0.9340 & 0.9190 & 0.9230  \\
QPT-ResNet50 \cite{zhao2023quality} & 0.8947 & 0.9141 & 0.8875 & 0.9109 & 0.9271 & 0.9413 & 0.9250 & \textbf{0.9279}\\ \hline
GMC-IQA & \textbf{0.9062} & \textbf{0.9225} & \textbf{0.9059} & \textbf{0.9182} & \textbf{0.9325} & \textbf{0.9471} & \textbf{0.9251} & 0.9267 \\ \hline
\end{tabular}
}
\caption{Comparison of NR-IQA models on four authentic IQA datasets. The highest scores are marked in black bold.}
\label{tab:individual_datasets_comparation}
\end{table*}

\noindent\textbf{B) Cross-dataset Evaluation.}
We compare the cross-dataset performance of our model with three NR-IQA models, namely HyperIQA \shortcite{su2020blindly}, TReS \shortcite{golestaneh2022no} and MANIQA \shortcite{yang2022maniqa}. 
To validate the generalization of our network architecture, we train a variant model \emph{MoNet-MSE}, which optimizes the MoNet only using the MSE loss and employs the same training configurations as our full model.
We randomly choose 80\% images of one dataset for training, and use another three datasets with all images for testing.
The results in \tbl{tab:srocc_cross_data} demonstrate the superiority of MoNet-MSE on multiple datasets, validating the robust generalization of the proposed MoNet architecture.
It can be seen that our full model GMC-IQA further improves the performance, although in some cases, its performance is slightly lower than that of MANIQA, such as a difference of 0.48\% and 0.59\% when trained on the SPAQ and tested on the LIVEC and Koniq. The results suggest that the proposed GMC loss can further enhance the generalization, enabling our model to extend its capabilities across a wider range of image distributions.
\begin{table}[h]
\centering
\resizebox{\columnwidth}{!}{
\begin{tabular}{cc|ccc|cc}
\hline
Training & Testing & HyperIQA & TReS &MANIQA & MoNet-MSE & GMC-IQA \\ 
\hline
 & Koniq & 0.7427 & 0.7307 & {\color[HTML]{0070C0} \textbf{0.7970}} & 0.7948 & \textbf{0.8015} \\
 & BID & 0.8708 & 0.8594 & 0.8759 & {\color[HTML]{0070C0} \textbf{0.8763}} & \textbf{0.8782} \\
\multirow{-3}{*}{LIVEC} & SPAQ & 0.8489 & 0.8655 & 0.8759 & {\color[HTML]{0070C0} \textbf{0.8763}} & \textbf{0.8773} \\ \hline
 & LIVEC & 0.7606 & 0.7704 & {\color[HTML]{0070C0} \textbf{0.8655}} & 0.8599 & \textbf{0.8677} \\
 & BID & 0.8018 & 0.8149 & {\color[HTML]{0070C0} \textbf{0.8571}} & 0.8529 & \textbf{0.8573} \\
\multirow{-3}{*}{Koniq} & SPAQ & 0.8354 & 0.8465 & \textbf{0.8908} & 0.8874 & {\color[HTML]{0070C0} \textbf{0.8880}} \\ \hline
 & LIVEC & 0.7717 & 0.7541 & 0.8328 & {\color[HTML]{0070C0} \textbf{0.8337}} & \textbf{0.8405} \\
 & Koniq & 0.6880 & 0.7071 & 0.7673 & \textbf{0.7819} & {\color[HTML]{0070C0} \textbf{0.7751}} \\
\multirow{-3}{*}{BID} & SPAQ & 0.8290 & {\color[HTML]{0070C0} \textbf{0.8390}} & 0.8363 & 0.8389 & \textbf{0.8416} \\ \hline
 & LIVEC & 0.7596 & 0.7814 & \textbf{0.8367} & {\color[HTML]{0070C0} \textbf{0.8358}} & 0.8327 \\
 & Koniq & 0.7336 & 0.7468 & \textbf{0.8249} & 0.8181 & {\color[HTML]{0070C0} \textbf{0.8200}} \\
\multirow{-3}{*}{SPAQ} & BID & 0.7755 & 0.7787 & 0.8048 & {\color[HTML]{0070C0} \textbf{0.8118}} & \textbf{0.8180} \\ \hline
\end{tabular}}
\caption{Cross-dataset evaluation on four authentic datasets in terms of SROCC. The highest score and the second highest score are marked in black bold and blue bold, respectively.}
\label{tab:srocc_cross_data}
\end{table}

\subsection{Performance Evaluation on the Loss Function} 
\noindent\textbf{A) Adapability Validation.}
To validate the adapability of the proposed GMC loss to various network architectures, we employ it to train three advanced NR-IQA models, namely HyperIQA \shortcite{su2020blindly}, MANIQA \shortcite{yang2022maniqa} and TeacherIQA\shortcite{chen2022teacher}. We also use it to train two versions of the proposed MoNet, which adopt ViT-Small and ViT-Base as the backbone, respectively. For comparison, we train these networks using the distance-based loss and the Norm loss \shortcite{li2020norm}. 
As shown in \tbl{tab:loss_comparation}, it is evident that all the networks obtain the highest scores on both the LIVEC and BID datasets when the GMC loss is used, proving the advantages of the GMC loss over the compared losses. 
Without any modification to the network architecture, the GMC loss can be easily integrated with existing IQA models and promote their performance.

\begin{table}[ht]
\resizebox{\columnwidth}{!}{
\begin{tabular}{c|ccc|ccc}
\hline
Dataset & \multicolumn{3}{c|}{LIVEC} & \multicolumn{3}{c}{BID} \\ \hline
Models & Dist & Norm & GMC & Dist & Norm & GMC \\ \hline
HyperIQA &0.8650 	&0.8591 	&\textbf{0.8741(+1.05\%)} & 0.8345 &	0.8444 & 	\textbf{0.8466(+1.45\%)}  \\
TeacherIQA & 0.8411 & 0.8523 &		\textbf{0.8529(+1.41\%)}  & 0.7663 & 0.7592 &	\textbf{0.7788(+1.63\%)} \\
MANIQA  & 0.8914 & 0.9025 & \textbf{0.9043(+1.45\%)}& 0.8755 & 0.8889 & \textbf{0.8868(+1.29\%)} \\ \hline
Ours(Small) & 0.8580 &	0.8705 &	\textbf{0.8709(+1.50\%)} & 0.8380 	& 0.8435 &	\textbf{0.8517(+1.63\%)}  \\ 
Ours(Base) & 0.8937 &	0.8982 &	\textbf{0.9062(+1.40\%)}&	0.8960 &	0.8966 &	\textbf{0.9059(+1.10\%) }\\ \hline
\end{tabular}
}
\caption{Evaluation on different losses in terms of SROCC.}
\label{tab:loss_comparation}
\end{table}

\begin{figure}[ht]
  \centering
  \hspace*{-7mm}
  \includegraphics[width=\columnwidth]{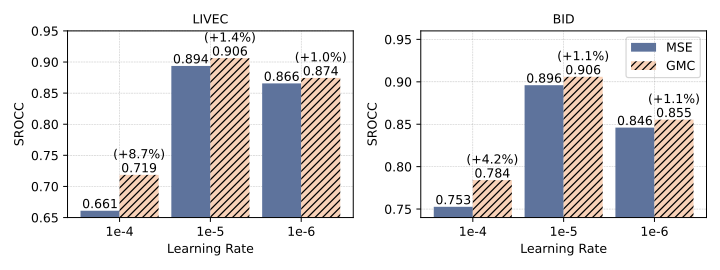}
  \caption{Comparison between the MSE loss and the GMC loss under different learning rates.}
  \label{fig:stable_comparison}
\end{figure}

\noindent\textbf{B) Sensitivity to Hyper-parameters.}
In \fig{fig:stable_comparison} we compare the performance of the MoNet trained with the MSE loss and the GMC loss under different learning rates (lr), including 1e-4, 1e-5 and 1e-6, on LIVEC and BID datasets. The results show that both losses achieve the highest scores when the lr is set to 1e-5 and perform the worst when the lr is set to 1e-4. Nevertheless, the GMC loss always outperforms the MSE loss. Especially when the lr is set to 1e-4, the GMC loss shows a more significant advantage with margins of 8.7\% and 4.2\% on the two datasets, respectively. The results indicate that the GMC loss is more robust to the setting of the lr, ensuring the training stability of the model.

\begin{figure}[ht]
  \centering
\includegraphics[width=\columnwidth]{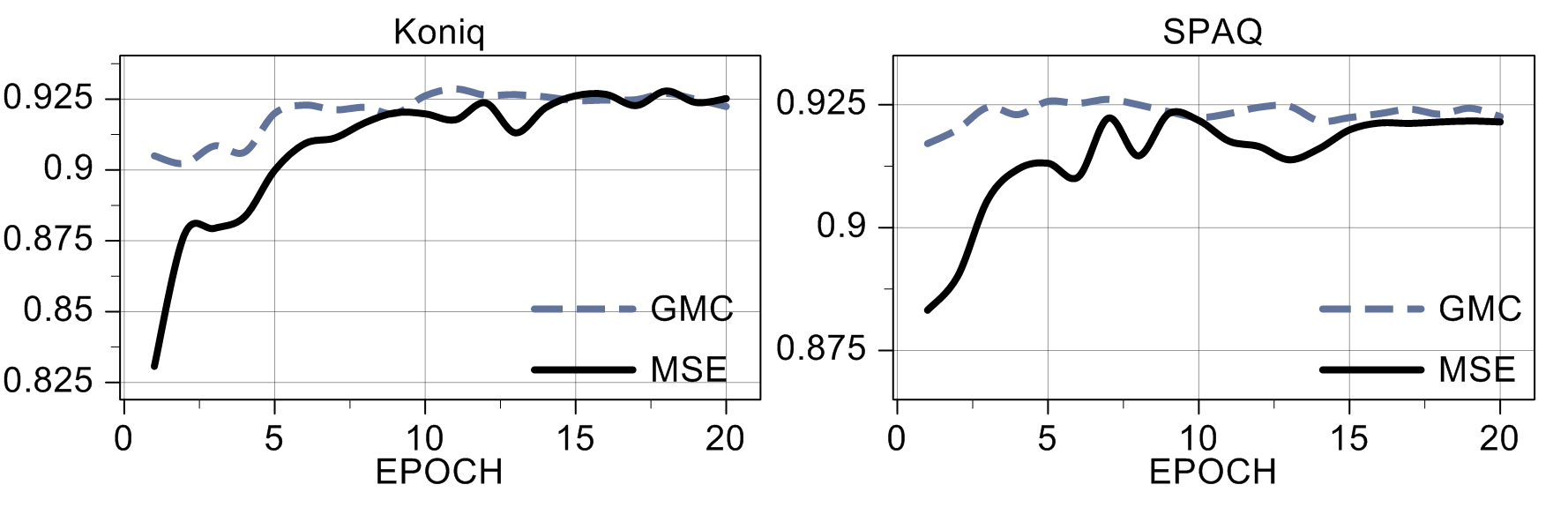}
  \caption{SROCC curves of the MoNet trained with the MSE and the GMC loss on the validation set of Koniq and SPAQ datasets.}
  \label{fig:learning_rate}
\end{figure}

\noindent\textbf{C) Convergence Comparison.}
\fig{fig:learning_rate} compares the SROCC curve of the MoNet trained with the MSE loss and the GMC loss on the validation set of the Koniq and SPAQ datasets. We can clearly see that the performance of the MoNet using the GMC loss surpasses 0.9 when it is trained with only one epoch. Moreover, the curve of the GMC loss is much smoother, demonstrating a faster convergence compared to the MSE loss. The results further validate the benefit of the GMC loss in enhancing the training stability, allowing us to train a model using fewer epochs.

\subsection{More Discussions and Analysis}

\noindent\textbf{A) Ablation Studies.}
To validate the effectiveness of the proposed key components, we train five variants of our model:
\romannumeral1) w/o MAL, which replaces the three MALs in the MoNet by three simple blocks, respectively, and each block consists of three convolutional layers;
\romannumeral2) w/o SGCC and \romannumeral3) w/o PGCC, which remove the SGCC loss and the PGCC loss from our total loss function defined in Eq. (\ref{eq:final_loss}), respectively; 
\romannumeral4) w/o GCC, which removes the two GCC losses and only adopts the MSE loss for training;
\romannumeral5) w/o queue, which removes previous data and only uses the current batch of data to compute the two GCC losses.
The results from \tbl{tab:ablation_study} reveal that the removal of any component degrades the model's performance. We can see that removing the MAL has the most remarkable decline in performance, validating the significance of the diverse opinion feature learning. In addition, removing the queue mechanism also shows a noticeable drop in performance, indicating that the utilization of previous data is beneficial to promoting the accuracy of the estimated metric scores. Moreover, removing both GCC losses demonstrates a more significant decrease than removing any one of them, suggesting that the joint optimization of the two GCC losses imposes stronger constraints on the global consistency.  
\begin{table}[ht]
\centering
\tiny
\resizebox{0.91\columnwidth}{!}{
\renewcommand\arraystretch{1}{
\begin{tabular}{c|cc|cc}
\hline
Dataset & \multicolumn{2}{c|}{LIVEC} & \multicolumn{2}{c}{BID}\\
\cline{2-5}
Model & SROCC & PLCC & SROCC & PLCC \\ \hline
w/o MAL & 0.8912 &	0.9109 &	0.8938 &	0.9078  \\ %\hline
w/o SGCC & 0.9017 &	0.9188   & 0.9010 &	0.9119\\
w/o PGCC & 0.9021 	&0.9183 	&0.9025 	&0.9158  \\
w/o GCC & 0.8937 &	0.9135  & 0.8960 &	0.9103  \\ 
w/o queue &  0.8962 & 0.9188 & 0.8936 & 0.9108  \\ 
\hline
Full model & \textbf{0.9062} & \textbf{0.9225} & \textbf{0.9059} & \textbf{0.9182}  \\ \hline
\end{tabular}
}}
\caption{Ablation studies on the critical components of our framework. The highest scores are marked in black bold.}
\label{tab:ablation_study}
\end{table}

\noindent\textbf{B) Discussion about the Length of the Queue.}
To investigate the effect of the queue length $K$ on the performance of our model, we re-train our model using different settings of $K$. Specifically, we set $K=r \times D$, where $r$ denotes the ratio of the training set size $D$, and $r$ is set to 20\%,  40\%, 60\%, 80\%. The results on the LIVEC, BID and SPAQ datasets are illustrated in \tbl{tab:queue_length_discussion}. It can be seen that as $K$ increases, the performance of our model first increases and then declines. This conforms our assumption that neither $K$ is too small nor too large is beneficial to the accuracy. 
\begin{table}[ht]
\resizebox{\columnwidth}{!}{
\begin{tabular}{c|cc|cc|cc}
\hline
Dataset & \multicolumn{2}{c|}{LIVEC} & \multicolumn{2}{c|}{BID} & \multicolumn{2}{c}{SPAQ} \\ \hline
Length Ratio & SROCC & PLCC & SROCC & PLCC & SROCC & PLCC \\ \hline
20\% & 0.9030 & 0.9183 & 0.9004 & 0.9141 & 0.9232 & 0.9256 \\
40\% & \textbf{0.9062} & 0.9194 & 0.9040 & 0.9159 & 0.9230 & 0.9251 \\
60\% & \textbf{0.9062} & \textbf{0.9225} & \textbf{0.9059} & \textbf{0.9182} & \textbf{0.9251} & \textbf{0.9267} \\
80\% & 0.9003 & 0.9198 & 0.9020 & 0.9145 & 0.9235 & 0.9266 \\ \hline
\end{tabular}
}
\caption{The impact of the queue length, which is set to different ratios of the training set size. The highest scores are marked in bold.}
\label{tab:queue_length_discussion}
\vspace{-0.2cm}
\end{table}

\begin{figure}[ht]
  \centering
\vspace{-0.1cm}
\setlength{\abovecaptionskip}{0.cm}
  \includegraphics[width=0.95\columnwidth]{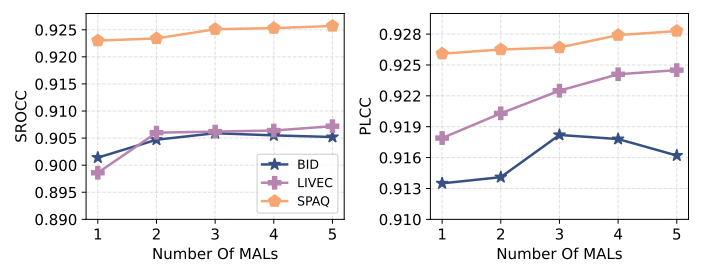}
  \caption{The impact of the MAL's number on the performance of our model on three datasets.}
  \label{fig:MAL_Num}
  \vspace{-0.2cm}
\end{figure}

\noindent \textbf{C) Discussion about the Number of the MAL.}
To explore the effect of the MAL's number $M$ on the performance of our model, 
we re-train our model using different settings of $M$ (1, 2, 3, 4 and 5).
The results on the BID, LIVEC and SPAQ datasets are illustrated in \fig{fig:MAL_Num}. We can see that with the increase of $M$, our model consistently demonstrates an improved performance on the LIVEC and SPAQ. This indicates that incorporating more MALs can benefit the performance, since more complementary contexts are learned. 
Additionally, we find that the scores on the BID initially experience an upward trend and subsequently decline. We speculate that this stems from the constraints imposed by limited training samples and single distortion type. These limitations weaken the model's capacity in capturing diverse attentions.

\begin{figure}[ht]
  \centering
  \includegraphics[width=0.5\textwidth]{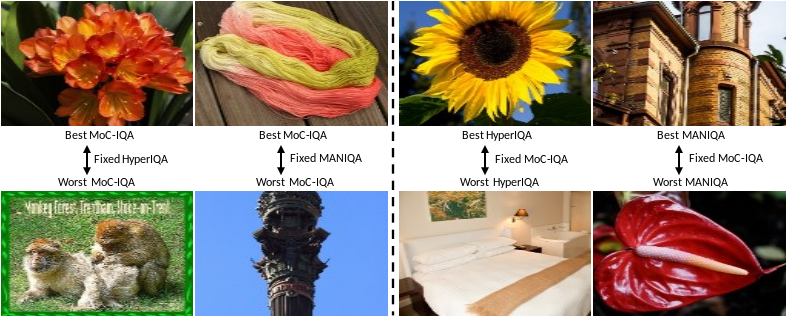}
  \caption{The gMAD competition against HyperIQA and MANIQA on the Koniq dataset.
  }
  \label{fig:gmad}
  \vspace{-0.3cm}
\end{figure}

\noindent \textbf{D) GMAD competition.}
In the gMAD \cite{ma2016group}, an attacker challenges a defender to select image pairs, where the defender perceives them as the same quality while the attacker perceives them differently. 
We conduct the gMAD with HyperIQA \cite{su2020blindly} and MANIQA \cite{yang2022maniqa} on the Koniq.
As shown in \fig{fig:gmad}, when our model acts as the attacker (leftmost two columns), it accurately recognizes the image pairs with noticeable quality differences while the defenders regard them with similar quality. 
Conversely, when our model acts as the defender (rightmost two columns), the attackers struggle to distinguish the quality difference of the images selected by our model.

\begin{figure}[ht]
  \centering
  \includegraphics[width=0.5\textwidth]{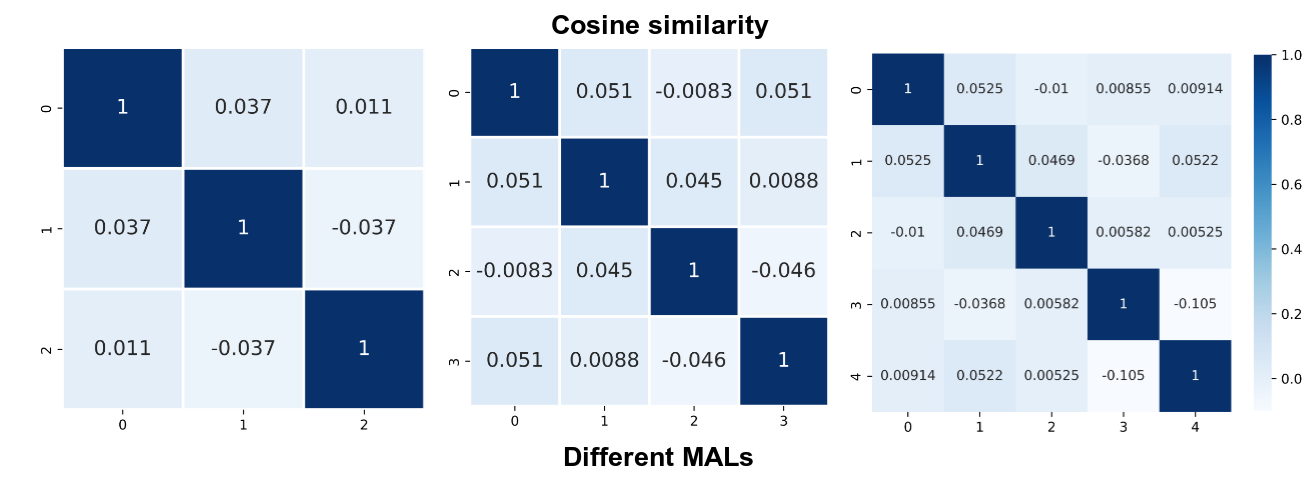}
  \caption{Cosine similarities between pairwise MALs. From left to right the number of MALs is set to 3, 4 and 5. }
  \label{fig:different_mals}
  \vspace{-0.4cm}
\end{figure}

\begin{figure}[ht]
  \centering
\setlength{\abovecaptionskip}{0.1cm}
  \includegraphics[width=\columnwidth]{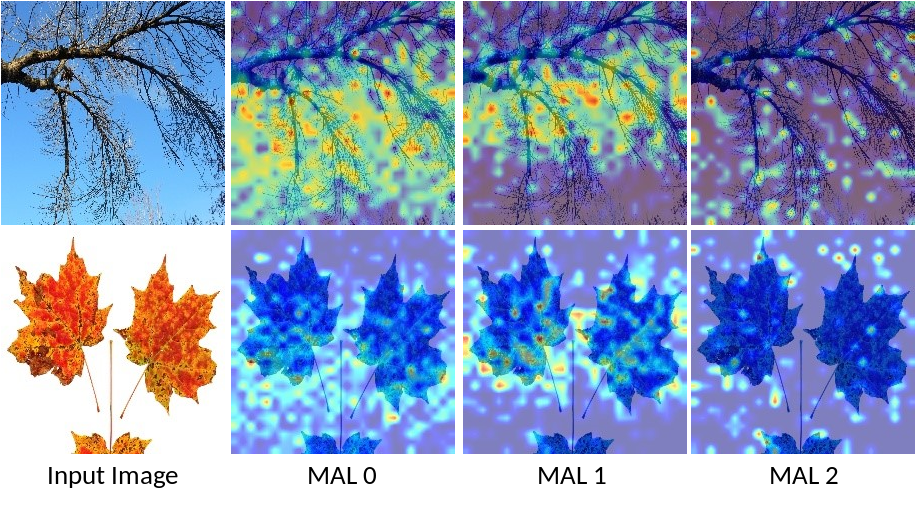}
  \caption{Attention maps produced by different MALs. The number of MALs is set to 3.
  }
  \label{fig:visual_mals}
  \vspace{-0.5cm}
\end{figure}

\noindent \textbf{E) Visualization Analysis on the MAL.}
To validate that the proposed MALs can learn diverse attentions, we compute the cosine similarity between the weights of each pairwise MALs and show it in \fig{fig:different_mals}. We see that all the similarity scores except those in the diagonal are extremely low, meaning that there exists little redundancy between each pairwise MALs. 
More intuitively, we visualize the output of different MALs in \fig{fig:visual_mals}. It can be observed that different MALs have distinct attention regions. For example, the first MAL pays more attention to global semantics, the second MAL mainly focuses on local salient regions, and the third MAL prefers the background. 
The examples show that each MAL effectively learns complementary opinion features.

\section{Conclusion}
In this paper, we develop a kind of GCC loss, which is combined with the MSE loss for jointly optimizing global-correlation consistency and mean-opinion consistency. Additionally, we develop a MoNet that aggregates various opinion features by capturing multi-view contexts, which effectively strengthens the robustness of the network architecture. We adopt the proposed loss to optimize the MoNet, achieving better performance, more stable training and faster convergence.
Extensive experimental results have demonstrated the superior accuracy and generalization achieved by our model on multiple authentic IQA datasets. Moreover, the proposed loss can be easily integrated with existing IQA models and further boost their performance.

\bibliographystyle{named}
\bibliography{ijcai24}

\end{document}